\pdfoutput=1

\documentclass{article}
\usepackage{ijcai24}

\usepackage{times}
\usepackage{soul}
\usepackage{url}
\usepackage[hidelinks]{hyperref}
\usepackage[utf8]{inputenc}
\usepackage[small]{caption}
\usepackage{graphicx}
\usepackage{amsmath}
\usepackage{amsthm}
\usepackage{booktabs}
\usepackage{algorithm}
\usepackage{algorithmic}
\usepackage[switch]{lineno}
\usepackage{amssymb}
\usepackage{multirow}
\usepackage{booktabs}
\usepackage{newclude}
\usepackage[hidelinks]{hyperref}
\usepackage[table,xcdraw]{xcolor}
\usepackage{color}
\usepackage{makecell}
\usepackage{stfloats}
\usepackage{url}

\hypersetup{
	colorlinks=true,
	dvipdfm=true,
	linkcolor=red,
	anchorcolor=blue,
	citecolor=green,
	urlcolor=magenta,
	pdfusetitle=true,
}

\title{LeMeViT: Efficient Vision Transformer with Learnable Meta Tokens for Remote Sensing Image Interpretation}

\author{
Wentao Jiang$^1$\and
Jing Zhang$^2$\and
Di Wang$^1$\and
Qiming Zhang$^2$\and
Zengmao Wang$^1$\footnote{Corresponding Author} \And
Bo Du$^1$\\
\affiliations
$^1$Wuhan University  $^2$The University of Sydney
\emails
\{jiang\_wentao, wangzengmao, dubo, d\_wang\}@whu.edu.cn,
jing.zhang1@sydney.edu.au, qzha2506@uni.sydney.edu.au
}

\begin{document}

\maketitle

\begin{abstract}

Due to spatial redundancy in remote sensing images, sparse tokens containing rich information are usually involved in self-attention (SA) to reduce the overall token numbers within the calculation, avoiding the high computational cost issue in Vision Transformers. However, such methods usually obtain sparse tokens by hand-crafted or parallel-unfriendly designs, posing a challenge to reach a better balance between efficiency and performance. Different from them, this paper proposes to use learnable meta tokens to formulate sparse tokens, which effectively learn key information meanwhile improving the inference speed. Technically, the meta tokens are first initialized from image tokens via cross-attention. Then, we propose Dual Cross-Attention (DCA) to promote information exchange between image tokens and meta tokens, where they serve as query and key (value) tokens alternatively in a dual-branch structure, significantly reducing the computational complexity compared to self-attention. By employing DCA in the early stages with dense visual tokens, we obtain the hierarchical architecture LeMeViT with various sizes. Experimental results in classification and dense prediction tasks show that LeMeViT has a significant $1.7 \times$ speedup, fewer parameters, and competitive performance compared to the baseline models, and achieves a better trade-off between efficiency and performance. The code is released at \url{https://github.com/ViTAE-Transformer/LeMeViT}.

\end{abstract}
\section{Introduction}

\begin{figure}[t]
   {\centering
   \includegraphics[width=0.5 \textwidth]{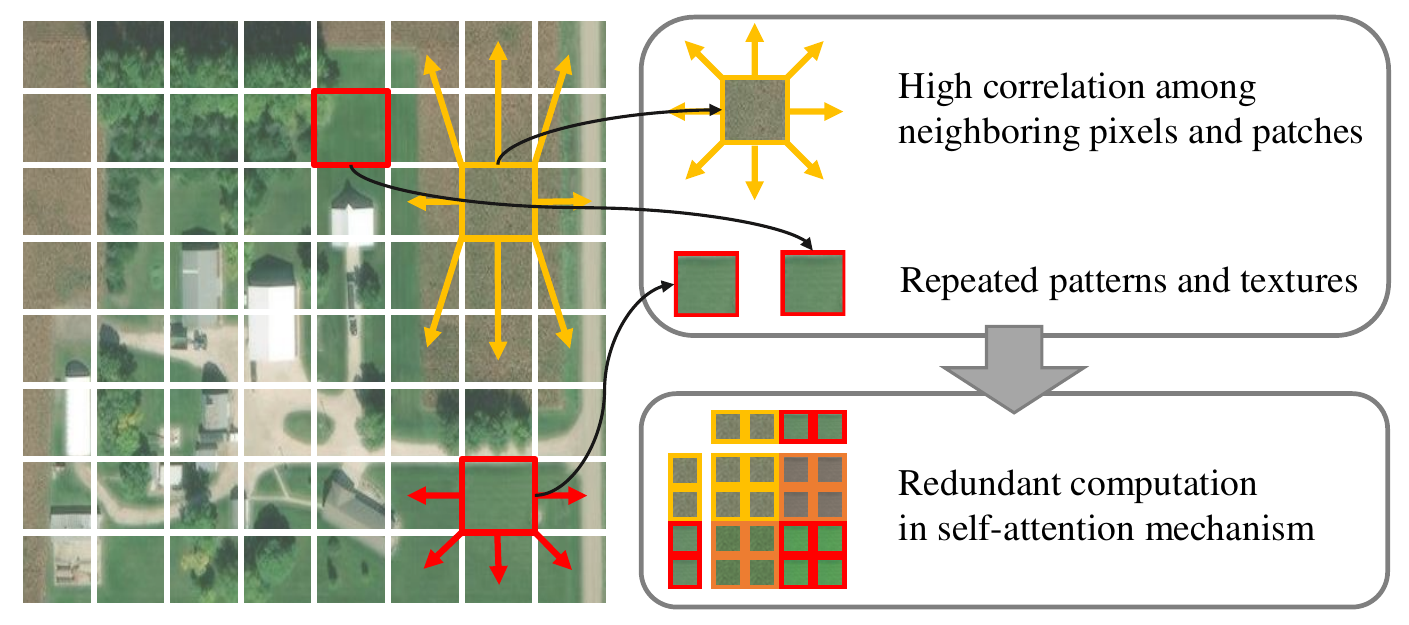}
   \caption{
Due to the high correlation between neighboring pixels and image patches, as well as the repetitive nature of textures and patterns in remote sensing images, there is a significant amount of spatial redundancy. This results in redundant computation in self-attention mechanism.
   \label{fig:motivation1}
   }
   }
\end{figure}

Since the remarkable success of migrating Transformer \cite{vaswani_attention_2017} from the field of natural language processing to the domain of computer vision, Vision Transformer (ViT) \cite{dosovitskiy_image_2021} has sparked significant interest in the community, highlighting great progress and advancements ~\cite{carion_end--end_2020, touvron_training_2021}. Several works \cite{xu_vitae_2021, zhang_2023_vitaev2} demonstrate ViT can model long-range dependency within visual information compared to traditional CNN networks with inherent inductive bias, unveiling its revolutionary potential in vision tasks including remote sensing image interpretation ~\cite{bazi_vision_2021,zhang_trs_2021,wang_advancing_2023}. 

However, due to the significant spatial redundancy in remote sensing images, ViT suffers from redundant computational overhead, as illustrated in Fig.~\ref{fig:motivation1}. The self-attention mechanism in ViT computes pairwise affinities between 
each two image patches regardless of how much useful information the tokens contain. Consequently, the `background' homogeneous tokens may contribute marginally to the informative feature representations but consume much compute load, hindering the whole model's efficiency.

To address this issue, some works discover the sparse representation of redundant image tokens \cite{chen_chasing_2021} in the natural image domain. These approaches use a shorter token sequence to represent the original image tokens and replace standard pairwise attention with cross-attention between image tokens and reduced tokens, decreasing the complexity of attention computation. For example, PVT \cite{wang_pyramid_2021} obtains reduced tokens through convolutional downsampling, Paca-ViT \cite{grainger_paca-vit_2023} uses data-driven weight parameters to cluster image tokens, while BiFormer \cite{zhu_biformer_2023} selects a small subset of more informative tokens from coarse to fine level. These methods rely on strong priors that may overlook useful image information, which leaves room for exploring more effective sparse representation. Besides, some methods employ parallel-unfriendly operators like clustering, slowing down computation and increasing memory access.

\begin{figure}[t]
   {\centering
   \includegraphics[width=0.5 \textwidth]{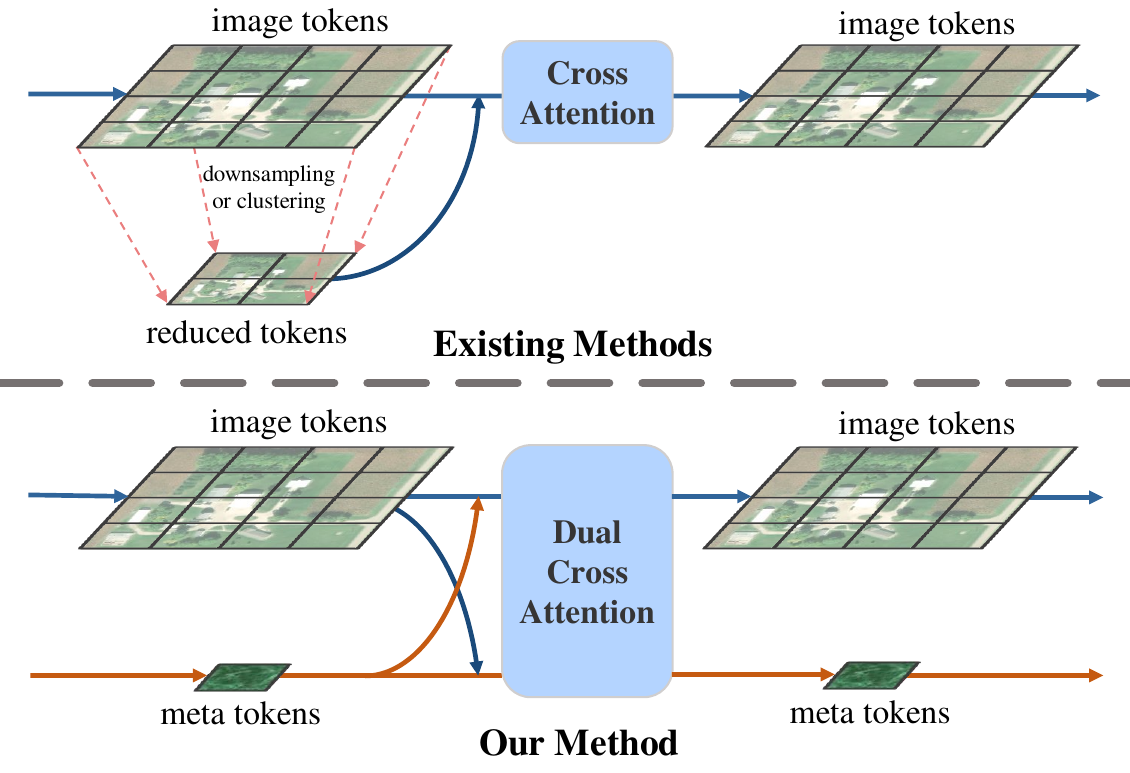}
   \caption{\textbf{Existing methods} commonly use downsampling or clustering to reduce the number of image tokens within the current block, which relies on strong priors or is parallel-computation unfriendly. \textbf{Our method} learns meta tokens to sparsely represent dense image tokens. Meta tokens exchange information with image tokens via the computationally efficient Dual Cross-Attention Block in an end-to-end way, promoting information flow stage-by-stage.
   \label{fig:motivation2}
   }
   }
\end{figure}

In this paper, we propose a \underline{Vi}sion \underline{T}ransformer with \underline{Le}arnable \underline{Me}ta Tokens called \textbf{LeMeViT}, aiming to leverage an extremely small number of learnable meta tokens to represent image tokens. 
Specifically, the meta tokens are first initialized from image tokens via cross-attention. Then, we propose Dual Cross-Attention (DCA) to promote information exchange between image tokens and meta tokens, where they serve as query and key (value) tokens alternatively in a dual-branch structure, significantly reducing the computational complexity from quadratic to linear compared to self-attention, as illustrated in Fig.~\ref{fig:motivation2}. By employing DCA in the early stages with dense visual tokens, we obtain the hierarchical architecture LeMeViT with various sizes.

\textbf{LeMeViT} has been intentionally designed to be hardware-friendly. Since modern GPUs excel at parallel computing and coalesced matrix operations, our model exclusively employs simple and dense operators, such as matrix multiplication, standard convolutions, and activation functions. The effective designs of the model architecture and utilization of hardware-friendly operators achieve nearly 1.7$\times$ speedup compared to the state-of-the-art (SOTA) model, \textit{e.g.}, ViTAE \cite{wang_empirical_2023}, while also improving performance. Additionally, the model can adapt to sequences of varying lengths, making it transferable to images of various resolutions, which is a common requirement in remote sensing tasks. Experimental results indicate that the model offers competitive performance while enjoying computational efficiency across multiple dense prediction tasks including semantic segmentation, object detection, and change detection.

Our contributions can be summarized as follows:
\begin{itemize}
    \item We propose a novel Transformer architecture called LeMeViT, which addresses the spatial redundancy in images via efficient architecture designs, achieving a better trade-off between efficiency and performance.
    \item We propose to learn sparse meta tokens to represent the dense image tokens and promote the information change between meta tokens and image tokens via a novel and computationally efficient DCA module. 
    \item Experiments on both natural images and remote sensing images demonstrate that LeMeViT achieves competitive performance compared to representative baseline models in both classification and dense prediction tasks. 
\end{itemize}

\section{Related Work}

\subsection{Vanilla Vision Transformer}
Transformer \cite{vaswani_attention_2017} quickly dominated the entire field of NLP since its inception and was later introduced into the realm of computer vision, which is known as Vision Transformer (ViT) \cite{dosovitskiy_image_2021}. DeiT \cite{touvron_training_2021} significantly alleviates the training difficulty of ViT, leading to a proliferation of ViT variants ~\cite{srinivas_bottleneck_2021,tu_maxvit_2022}, establishing a burgeoning and popular domain. ViT brings new vitality to vision tasks by modeling long-range dependencies. However, the significant computational complexity of the vanilla Transformer has remained a substantial challenge in practical usage, stemming in part from the quadratic complexity of its self-attention mechanism. Addressing this challenge become a hot topic of research.

\subsection{Efficient Vision Transformer}
A considerable amount of work is currently dedicated to reducing the complexity of self-attention, with mainstream approaches including sparse attention and token sparse representation. Sparse attention aims to reduce the connections between tokens, while token sparse representation aims to represent the image using fewer tokens.

One pattern of sparse attention, specifically local attention, has garnered significant interest following the success of the Swin Transformer~\cite{liu_swin_2021}, inspiring numerous works~\cite{zhang_vsa_2022, zhang_vision_2024}. 
Additionally, some works~\cite{liu_dynamic_2022} explore other forms of sparse attention. For instance, KVT~\cite{wang_kvt_2022} selects only the top-k similar keys for every query in attention. QuadTree Attention~\cite{tang_quadtree_2022} draws on the method of quadtree segmentation to partition tokens into square blocks, facilitating attention at different granularities. 

The earliest work utilizing token sparse representation may be PVT \cite{wang_pyramid_2021}, which reduces the number of keys and values through convolutional downsampling. CrossViT \cite{chen_crossvit_2021} utilizes image patches of larger size to reduce the number of tokens. Deformable Attention \cite{xia_vision_2022} employs learnable sampling points to sample tokens from image features. 

Some other methods, such as token pruning/merging \cite{kong_spvit_2022}, remove or combine certain tokens using score functions. A typical token merging approach like ToMe \cite{bolya_token_2023-1} has shown excellent results when applied to Stable Diffusion \cite{bolya_token_2023}.

\subsection{ViT for Remote Sensing}
ViT's excellent modeling capability has found wide applications in the field of remote sensing. Numerous endeavors have attempted to incorporate specific characteristics of remote sensing images into ViT, making it more applicable in this domain \cite{zhang_trs_2021, wang_vit-based_2022, deng_when_2022}. These efforts are confined to specific tasks. Recently, RSP \cite{wang_empirical_2023} adopted a remote sensing pre-training approach to train a foundational model for the remote sensing domain. It achieves state-of-the-art performance across various downstream tasks. However, this ViT-based model also suffers from a serious computational burden, making it an obstacle to practical deployment. Therefore, addressing this issue has become a pressing priority.

\begin{figure*}[t]
   {\centering
   \includegraphics[width=0.88\textwidth]{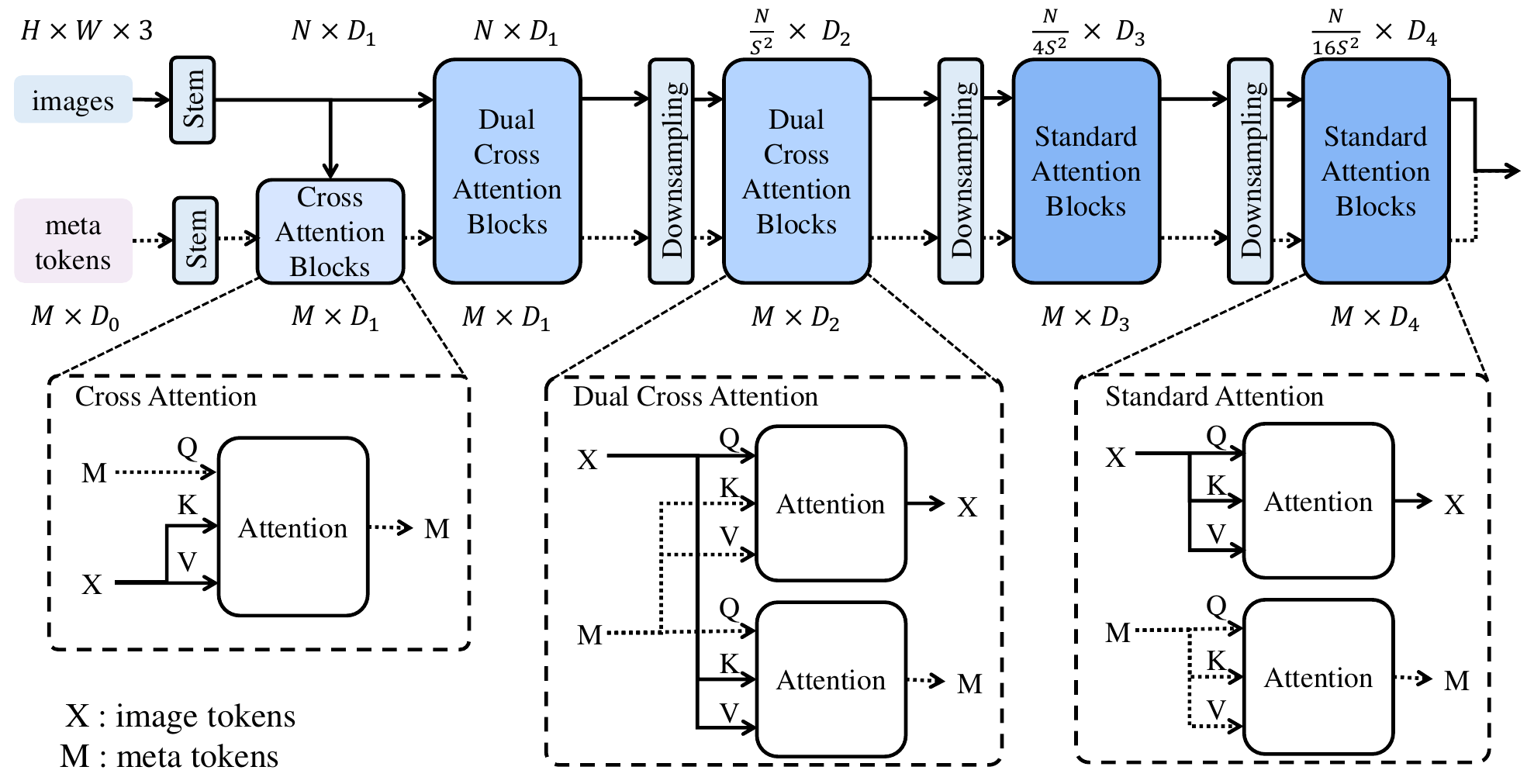}
   \caption{
      \textbf{The Overall Architecture of LeMeViT.} LeMeViT consists of three different attention blocks, arranged from left to right as Cross Attention Block, Dual Cross Attention Block, and Standard Attention Block. Specific details of attention computation method are provided. 
   \label{framework}
   }}
\end{figure*}

\section{Method}

\subsection{Overview and Preliminaries}
The overview of the proposed LeMeViT architecture is illustrated in Fig.~\ref{framework}. As depicted, LeMeViT follows a typical hierarchical ViT structure, with four stages connected by downsampling layers. As the stages get deeper, the spatial size of image features gradually reduces while feature dimensionality expands.The core components of the model are Meta Tokens and the DCA block. The meta tokens are first initialized from image tokens via cross-attention. Then, DCA is employed in the early stages to promote information exchange between image tokens and meta tokens, where they serve as query and key (value) tokens alternatively in a dual-branch structure. In the later stages, standard attention blocks based on self-attention are used.

In this paper, image tokens are represented as $\mathcal{X} \in \mathbb{R}^{N \times C}$ and meta tokens are represented as $\mathcal{M} \in \mathbb{R}^{M \times C}$, where $C$ represents token dimension, $N$ and $M$ represents the number of image tokens and meta tokens, respectively. $N$ varies across different stages, with $N \gg M$ in the early stages. Additionally, we employ $D_1$, $D_2$, $D_3$, and $D_4$ to signify the token dimensions across different stages.

Due to the extensive use of scaled dot-product attention in our model, we provide its formal definition here:
\begin{equation}
    \text{Attention}(Q, K, V) = \text{softmax}\left(\frac{QK^{\top}}{\sqrt{d_k}}\right) V,
\end{equation}
where queries $Q \in \mathbb{R}^{N_1 \times C} $, keys and values $K,V \in \mathbb{R}^{N_2 \times C} $. Scalar factor $\sqrt{d_k}$ is introduced to avoid gradient vanishing, where $d_k$ is normally assigned token dimension $C$. For maintaining entropy-invariance, we use $\frac{\log{N_1}}{\log{N_2}} \sqrt{C} $ as scalar factor in cross-attention instead \cite{chiang-cholak-2022-overcoming}.

\subsection{Key Components in LeMeViT}
Fig.~\ref{framework} shows all the components of LeMeViT, including learnable meta tokens, stem blocks, downsampling layers, and three types of attention blocks. Subsequently, we will introduce the detailed structures. 

\textbf{Learnable Meta Tokens.} Meta tokens are a set of learnable tensors updated via information exchange with image tokens. They can be analogized to learnable queries in DETR \cite{carion_end--end_2020}, although their learning method and function differ. 
After training, initial meta tokens are fixed, but they continue to update by interacting with image tokens. They serve as the model input alongside the image tokens. Initially, the shape of meta tokens is $M \times D_0$. Their dimensions expand as image tokens, but their length remains $M$. Based on empirical analysis, we set $M$ to 16 in our experiments, as validated in the ablation studies (Sec.~\ref{ablation}).

\textbf{Stem and Downsampling Layers.} The Stem block divides the input image into patches and embeds them into tokens. We employ an overlapping patch embedding technique. Specifically, we implement this block using two $3\times3$ convolutions with a stride of 2 and padding of 1. The convolutional windows slide in an overlapping manner, and after two layers, the image is precisely divided into tokens, each of which corresponds to a patch of size $4\times4$. To align with the dimensions of image tokens, we introduce an extra Stem block for meta tokens, which consists of two MLP layers. After passing through the Stem block, the image is transformed into $N$ tokens, and we have $N = \frac{H}{4} \times \frac{W}{4}$. Both the image tokens and meta tokens share the feature dimension $D_1$. The downsampling layer similarly adopts an overlapping patch embedding approach but employs only one convolution to achieve a downsampling ratio of 2.

Then, three distinct attention blocks are used in LeMeViT, including the Cross Attention (CA) block, Dual Cross Attention (DCA) block, and Standard Attention (SA) block. They share similar structures, involving Conditional Positional Encodings (CPE), LayerNorm (LN), Attention, Feed Forward Network (FFN), and residual connections. Their only difference lies in the Attention layer. Notably, meta tokens and image tokens share the same FFN for parameter efficiency.

\textbf{Cross Attention Block.} CA block is employed to learn meta tokens from image tokens. Due to the considerable gap between the initial meta tokens and image tokens, directly using meta tokens as keys and values to update the image tokens might lead to the collapse of image features and information loss. Therefore, CA is designed to only update meta tokens. It employs a cross-attention mechanism, where query is the projection of meta tokens, and key and value are projections of image tokens. Its formulation can be described as follows:
\begin{equation}
    \mathcal{M} \Leftarrow \text{Attention}(\mathcal{M}_Q, \mathcal{X}_K, \mathcal{X}_V),
\end{equation}
where $\mathcal{M}_Q$ denotes query projection of meta tokens and $\mathcal{X}_K, \mathcal{X}_V$ denote key and value projections of image tokens.

\textbf{Dual Cross-Attention Block.} DCA block is the core component for enhancing computational efficiency. It replaces the pairwise self-attention among image tokens with two cross-attention between image tokens and meta tokens, reducing the computational complexity from quadratic $\mathcal{O}(N^2)$ to linear $\mathcal{O}(2MN)$. Considering that $M \ll N$, the efficiency improvement is notably evident. Meanwhile, it retains strong representation capabilities. Unlike other cross-attention strategies \cite{wang_pvt_2022,zhu_biformer_2023} that reduce image tokens explicitly, DCA implicitly preserves most of the image information from all stages through meta tokens. Within the DCA block, image tokens fuse global information held by meta tokens via cross-attention while aggregating local information of each patch into meta tokens via another cross-attention. Specifically, image tokens and meta tokens serve as each other's query and key/value, which can be formulated as:
\begin{equation}
    \mathcal{X} \Leftarrow \text{Attention}(\mathcal{X}_Q, \mathcal{M}_K, \mathcal{M}_V).
\end{equation}
\begin{equation}
    \mathcal{M} \Leftarrow \text{Attention}(\mathcal{M}_Q, \mathcal{X}_K, \mathcal{X}_V),
\end{equation}

\textbf{Standard Attention Block.} In the last two stages, we adopted the standard attention mechanism for the trade-off between efficiency and performance. In hierarchical ViTs, the number of image tokens decreases as the stages deepen. Consequently, the assumption of $M  \ll N$ may no longer hold in the final two stages. Additionally, due to the increased dimensions, computational and parameter overhead caused by projection layers becomes substantial. Therefore, DCA may not be as efficient as in the first two stages, making standard self-attention a preferable choice. Specifically, image tokens and meta tokens perform self-attention individually.

Finally, image tokens and meta tokens are processed by global average pooling separately and then being added together for the classification prediction.
Additionally, only image tokens from each stage, which have different scales, are used to perform dense prediction tasks.

\subsection{Details of Architecture Design}
Based on the overall architecture, we devise three model variants of different sizes, \textit{i.e.}, Tiny, Small, and Base. We tailored these sizes by adjusting the number of blocks and dimensions of features in each stage, as listed in Table \ref{arch}. Other configurations are shared between all variants. We set each head dimension of attention to 32, the MLP expansion rate to 4, and the conditional positional encoding kernel size to 3. The length of meta tokens is set to 16.

\begin{table}[h]
\centering
\setlength{\tabcolsep}{3pt}
\begin{tabular}{@{}c|ccccc|cccc@{}}
\toprule
\multirow{2}{*}{Version} & \multicolumn{5}{c|}{Blocks} & \multicolumn{4}{c}{Dims} \\ \cmidrule(l){2-10} 
                         & $S_0$  & $S_1$  & $S_2$  & $S_3$  & $S_4$  & $D_1$  & $D_2$   & $D_3$   & $D_4$   \\ \midrule
LeMeViT-Tiny                     & 1   & 2   & 2   & 8   & 2   & 64  & 128  & 192  & 320  \\
LeMeViT-Small                    & 1   & 2   & 2   & 6   & 2   & 96  & 192  & 320  & 384  \\
LeMeViT-Base                     & 2   & 4   & 4   & 18  & 4   & 96  & 192  & 384  & 512  \\ \bottomrule
\end{tabular}
   \caption{ \textbf{Architecture details of different LeMeViT variants.} $S_0$ to $S_4$ represent the number of blocks in the CA stage ($S_0$), two DCA stages ($S_1$ and $S_2$), and two SA stages ($S_3$ and $S_4$). $D_1$ to $D_4$ signify the dimensions of features in each stage, as shown in Fig.~\ref{framework}.
   \label{arch}}
\end{table}

\subsection{Computational Complexity Analysis} \label{computation complexity analysis}
We primarily analyze the computational complexity of DCA, the core module responsible for LeMeViT's efficiency improvement. We compute the complexity across the projection layer, attention layer, and FFN layer, forming an entire DCA block. We assume that the shape of image tokens is $N \times D$, the shape of meta tokens is $M \times D$, and the MLP expansion rate in FFN is $E$. The results are shown in Table~\ref{total complexity}.

\begin{table}[H]
\centering
\begin{tabular}{@{}c|l|l@{}}
\toprule
           & \multicolumn{1}{c|}{Dual Cross-Attention}     & \multicolumn{1}{c}{Standard Attention} \\ \midrule
Projection & $4ND^2+4MD^2$ & $4ND^2$                 \\
Attention  & $2NMD$                                          & $2N^2D$                                   \\
FFN        & $2E(N+M)D^2$                                     & $2E(N+M)D^2$                              \\  \bottomrule

\end{tabular}

\vspace{7pt}

\begin{tabular}{@{}c|l@{}}
\toprule
Standard Attention   & $(2E+4)ND^2+2N^2D$               \\
Dual Cross-Attention & $(2E+4)(N+M)D^2+2NMD$           \\ \bottomrule
\end{tabular}

   \caption{\textbf{Above:} Computation complexity of specific layers in DCA and standard attention. \textbf{Below:} Total computational complexity of the two attention blocks.
   \label{total complexity}}
\end{table}

Based on the results in Table~\ref{total complexity}, we can draw the following conclusion. The computational complexity of DCA is linear regarding $N$, which is significantly lower than the quadratic complexity of the standard attention. Specifically, for our three model variants, the computational complexity is reduced by about 10$\times$ compared to using standard attention, \textit{e.g.}, given an image size of 224, the first DCA block in the tiny/small/base model has 0.16/0.36/0.36 GFLOPs while the standard attention has 1.41/2.24/2.24 GFLOPs.
Experiments in Supplementary Material demonstrate that DCA achieve notably higher inference speed than SA.

\section{Experiments}

\begin{table}[]
\raggedright
\setlength{\tabcolsep}{2.2pt}
\begin{tabular}{@{}lcccc@{}}
\toprule
\textbf{Model}                                                      & \textbf{\begin{tabular}[c]{@{}c@{}}Infer $\uparrow$ \\ (img/sec)\end{tabular}} & \textbf{\begin{tabular}[c]{@{}c@{}}Params $\downarrow$\\ (M)\end{tabular}} & \textbf{\begin{tabular}[c]{@{}c@{}}MACs $\downarrow$\\ (G)\end{tabular}} & \textbf{\begin{tabular}[c]{@{}c@{}}Acc@1 $\uparrow$\\ (\%)\end{tabular}} \\ \midrule
\multicolumn{1}{l|}{PVTv2-b1}                                       & \multicolumn{1}{c|}{4897.43}                                       & 14.01                                                         & \multicolumn{1}{c|}{2.03}                                   & 78.70                                                         \\
\multicolumn{1}{l|}{MobileViTv2}                                    & \multicolumn{1}{c|}{5162.10}                                       & 4.90                                                          & \multicolumn{1}{c|}{1.41}                                   & 78.10                                                         \\
\multicolumn{1}{l|}{Efficientformerv2}                              & \multicolumn{1}{c|}{1617.41}                                       & 6.19                                                          & \multicolumn{1}{c|}{0.63}                                   & 79.00                                                         \\
\rowcolor[HTML]{D8D8D8} 
\multicolumn{1}{l|}{\cellcolor[HTML]{D8D8D8}\textbf{LeMeViT-tiny}}  & \multicolumn{1}{c|}{\cellcolor[HTML]{D8D8D8}\textbf{5316.58}}      & 8.64                                                          & \multicolumn{1}{c|}{\cellcolor[HTML]{D8D8D8}1.78}           & \textbf{79.07}                                                         \\ \midrule
\multicolumn{1}{l|}{Swin-tiny}                                      & \multicolumn{1}{c|}{2872.48}                                       & 28.29                                                         & \multicolumn{1}{c|}{4.35}                                   & 81.30                                                         \\
\multicolumn{1}{l|}{PVTv2-b2}                                       & \multicolumn{1}{c|}{2866.77}                                       & 25.36                                                         & \multicolumn{1}{c|}{3.88}                                   & 82.00                                                         \\
\multicolumn{1}{l|}{FLatten-Swin-T}                                 & \multicolumn{1}{c|}{1918.84}                                       & 28.50                                                         & \multicolumn{1}{c|}{4.39}                                   & \textbf{82.10}                                                         \\
\multicolumn{1}{l|}{FLatten-PVT-S}                                  & \multicolumn{1}{c|}{1863.83}                                       & 24.72                                                         & \multicolumn{1}{c|}{3.70}                                   & 81.70                                                         \\
\multicolumn{1}{l|}{PacaViT-tiny}                                   & \multicolumn{1}{c|}{2157.28}                                       & 12.20                                                         & \multicolumn{1}{c|}{3.10}                                   & 80.63                                                         \\
\multicolumn{1}{l|}{BiFormer-tiny}                                  & \multicolumn{1}{c|}{2889.70}                                       & 13.14                                                         & \multicolumn{1}{c|}{2.20}                                   & 81.40                                                         \\
\rowcolor[HTML]{D8D8D8} 
\multicolumn{1}{l|}{\cellcolor[HTML]{D8D8D8}\textbf{LeMeViT-small}} & \multicolumn{1}{c|}{\cellcolor[HTML]{D8D8D8}\textbf{3608.12}}      & 16.40                                                         & \multicolumn{1}{c|}{\cellcolor[HTML]{D8D8D8}3.74}           & 81.88                                                         \\ \midrule
\multicolumn{1}{l|}{Swin-small}                                     & \multicolumn{1}{c|}{1717.31}                                       & 49.61                                                         & \multicolumn{1}{c|}{8.51}                                   & 83.00                                                         \\
\multicolumn{1}{l|}{Swin-base}                                      & \multicolumn{1}{c|}{1215.39}                                       & 87.77                                                         & \multicolumn{1}{c|}{15.13}                                  & 83.30                                                         \\
\multicolumn{1}{l|}{PVTv2-b4}                                       & \multicolumn{1}{c|}{1494.79}                                       & 62.56                                                         & \multicolumn{1}{c|}{9.79}                                   & 83.60                                                         \\
\multicolumn{1}{l|}{CrossViT-base}                                  & \multicolumn{1}{c|}{\textbf{1911.69}}                                       & 105.03                                                        & \multicolumn{1}{c|}{20.10}                                  & 82.20                                                         \\
\multicolumn{1}{l|}{PacaViT-base}                                   & \multicolumn{1}{c|}{927.16}                                        & 46.91                                                         & \multicolumn{1}{c|}{9.26}                                   & 83.96                                                         \\
\multicolumn{1}{l|}{BiFormer-base}                                  & \multicolumn{1}{c|}{799.07}                                        & 56.80                                                         & \multicolumn{1}{c|}{9.32}                                   & 84.30                                                         \\
\rowcolor[HTML]{D8D8D8} 
\multicolumn{1}{l|}{\cellcolor[HTML]{D8D8D8}\textbf{LeMeViT-base}}  & \multicolumn{1}{c|}{\cellcolor[HTML]{D8D8D8}1482.70}                                                  & 53.10                                                         & \multicolumn{1}{c|}{\cellcolor[HTML]{D8D8D8}11.06}                                                       & \textbf{84.35}                                                         \\ \bottomrule
\end{tabular}
   \caption{Comparison of different models on ImageNet-1K.
   \label{tab:imagenet}
   }
\end{table}

To assess the efficiency and performance of our model, we first evaluate it on the ImageNet-1K dataset for image classification, comparing it against other efficient ViTs in Sec.~\ref{imagenet}. Then, we conduct a series of experiments in remote sensing tasks compared to the representative Swin Transformer~\cite{liu_swin_2021} and SOTA ViTAE~\cite{xu_vitae_2021}. Specifically, we pre-train the model on the MillionAID~\cite{long_creating_2021} dataset (Sec.~\ref{millionaid}) and then transfer it to downstream tasks including object detection, semantic segmentation, and change detection (Sec.~\ref{downstream}). Additionally, we conduct an ablation study to validate the setting of the length of meta tokens in Sec.~\ref{ablation}. Finally, we visualize and analyze the attention map in Sec.~\ref{visualization}.

\begin{table*}[t]
\centering
\renewcommand{\arraystretch}{1.1}
\setlength{\tabcolsep}{4.2pt}

\begin{tabular}{l!{\vrule width1.2pt}cc!{\vrule width1.2pt}cc!{\vrule width1.2pt}cc!{\vrule width1.2pt}cc}
   \bottomrule[1.5pt]
\multirow{2}{*}{Model} & \multicolumn{2}{c!{\vrule width1.2pt}}{\textbf{Throughputs (img/sec) $\uparrow$}}     & \multicolumn{2}{c!{\vrule width1.2pt}}{\textbf{Memory Usage (GB) $\downarrow$}}          & \multirow{2}{*}{\textbf{\begin{tabular}[c]{@{}c@{}}Params $\downarrow$\\ (M)\end{tabular}}} & \multirow{2}{*}{\textbf{\begin{tabular}[c]{@{}c@{}}MACs $\downarrow$\\ (G)\end{tabular}}} & \multirow{2}{*}{\textbf{\begin{tabular}[c]{@{}c@{}}Acc@1 $\uparrow$\\ (\%)\end{tabular}}} & \multirow{2}{*}{\textbf{\begin{tabular}[c]{@{}c@{}}Acc@5 $\uparrow$\\ (\%)\end{tabular}}} \\ \cline{2-5}
                       & \multicolumn{1}{c}{\textbf{Train}} & \textbf{Infer}   & \multicolumn{1}{c}{\textbf{Train}} & \textbf{Infer} &                                                                                &                                                                              &                                                                                &                                                                                \\ \Xhline{1pt}
Swin-tiny              & 872.85                              & 2874.62          & 6.14                                & 1.87           & 27.56                                                                          & 4.35                                                                         & 98.42 (98.59)                                                                  & 99.87 (99.88)                                                                  \\
ViTAEv2-small          & 624.09                              & 1847.65          & 8.44                                & 1.70           & 18.87                                                                          & 5.48                                                                         & 98.97                                                                          & 99.88                                                                          \\ \Xhline{0.1pt}
LeMeViT-tiny           & 1389.33                             & 5327.47          & 3.91                                & 1.36          & 8.33                                                                           & 1.78                                                                         & 98.80                                                                          & 99.82                                                                          \\

\multicolumn{1}{l!{\vrule width1.2pt}}{LeMeViT-small} & 968.59                     & 3612.68 & 5.33                       & 1.69  & 16.04                                                                 & 3.74                                                                & 99.00                                                                 & \textbf{99.90}                                                                 \\
LeMeViT-base           & 408.92                              & 1484.09          & 11.08                               & 1.91           & 52.61                                                                          & 11.06                                                                        & \textbf{99.17}                                                                          & 99.88                                                                          \\ \toprule[1.5pt]
\end{tabular}
   \caption{ Results of Scene Recognition on MillionAID. The results in the parentheses denote the results of Swin-tiny trained for 300 epochs, while the others are based on models trained for 100 epochs.
   \label{tab:millionaid}
   }
\end{table*}

\begin{figure}[t]
   {\centering
   \vspace{-2pt} 
   \includegraphics[width=0.48 \textwidth]{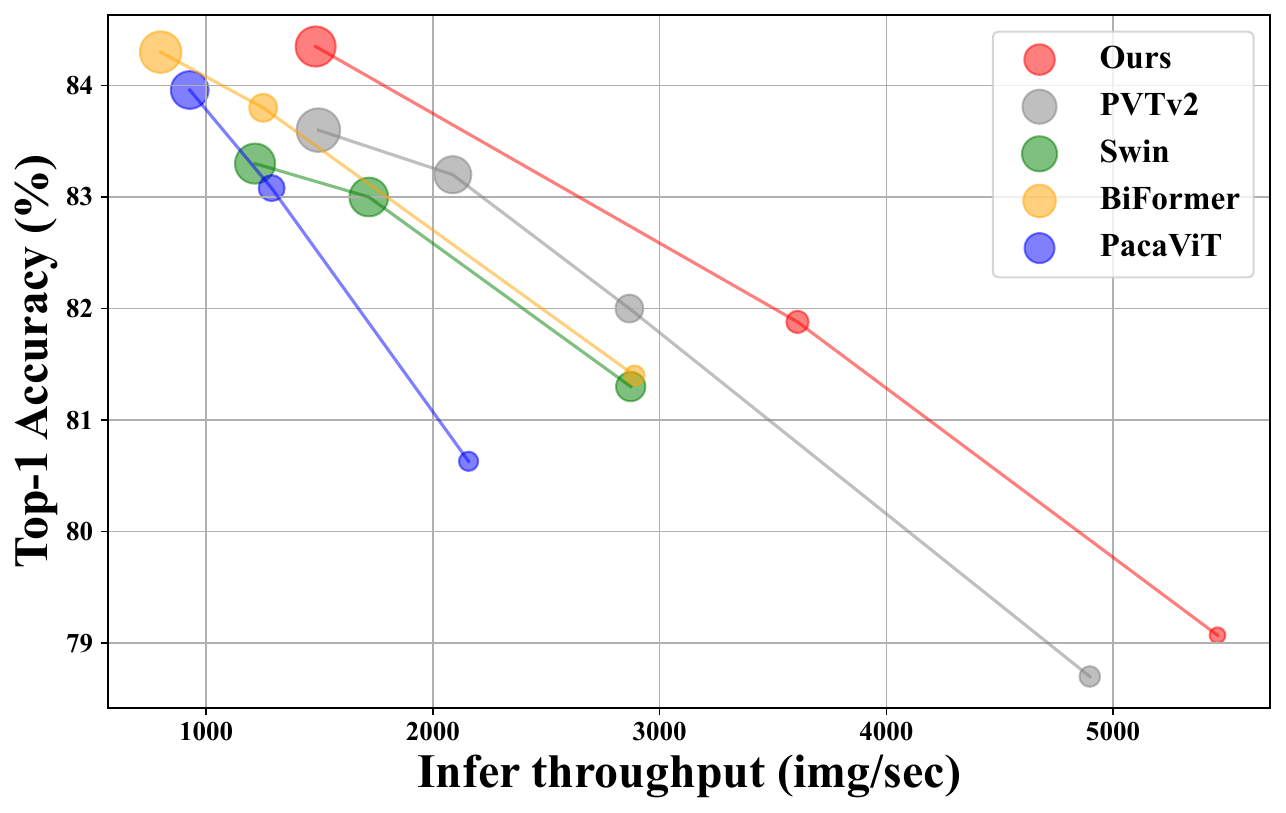}
   \caption{ Visualization of comparison between different models. The size of scatter represents the parameter count of the model.
   \label{fig:efficientcy}
   }
   }
\end{figure}

\subsection{Image Classification on ImageNet-1K} \label{imagenet}
We first conduct image classification experiments on the ImageNet-1K benchmark. We train the three variants of our model and compare them with other efficient ViTs of different sizes. These representative methods include Swin Transformer \cite{liu_swin_2021}, PVTv2 \cite{wang_pvt_2022}, CrossViT \cite{chen_crossvit_2021}, MobileViTv2 \cite{mehta_mobilevit_2022}, Efficientformerv2 \cite{li_rethinking_2023}, Flatten Attention \cite{han_flatten_2023}, Paca-ViT \cite{grainger_paca-vit_2023}, and BiFormer \cite{zhu_biformer_2023}. For a fair comparison, all our three models are trained using the same settings following DeiT, as mentioned below.

\textbf{Metrics.} We employ efficiency and classification performance metrics for evaluation, where efficiency metrics include throughput, parameter count, and MACs, while top-1 accuracy serves as the classification performance metric. However, MACs may have a low correlation with actual inference latency, due to the fact that MACs can not reflect memory access efficiency. Therefore, we primarily use throughput as the efficiency metric.

\textbf{Experiment Details.} The models are implemented by PyTorch and trained 300 epochs from scratch. 
We employ AdamW as the optimizer, and apply a cosine decay learning rate schedule. The complete hyper-parameters for learning strategy and data augmentation techniques are provided in the Supplementary Material.
We apply stochastic depth with 0.15 probability for our models. 
We set different batch sizes, \textit{i.e.}, 400, 256, and 64, for Tiny, Small, and Base model variants, respectively. We train the models on four Nvidia RTX 4090 GPUs and test on one. We report the top-1 accuracy on the ImageNet-1K validation set. For a fair comparison, we use the same TIMM benchmark tool to test the throughput, parameter count, and MACs of our models and different models in the same environment.

\textbf{Results.} Quantitative results are listed in Table~\ref{tab:imagenet}, and plotted in Fig.~\ref{fig:efficientcy}. The results show that our LeMeViT achieves the best trade-off in efficiency and performance among all the comparison methods, which can be easily observed from Fig.~\ref{fig:efficientcy}. For quantitative analysis, compared to the most competitive PVTv2 \cite{wang_pvt_2022}, LeMeViT-Base obtains 0.75\% better accuracy than PVTv2-b4, with similar throughput and fewer parameters. Compared to BiFormer-Base \cite{zhu_biformer_2023}, LeMeViT-Base achieves 1.85$\times$ speedup and slight accuracy increase. Compared to Swin Transformer and Paca-ViT, our model outperforms them both in throughput and accuracy. Taking into account the trade-off between accuracy, throughput, and parameter count, our model generally outperforms other efficient ViT models.

\subsection{Remote Sensing Scene Recognition} \label{millionaid}
After evaluating LeMeViT on natural image classification, we apply the model to the remote sensing domain. Following the remote sensing model RSP \cite{wang_empirical_2023}, we first pre-train our model on the MillionAID dataset for aerial scene recognition. Most experimental details remain consistent with the classification task, with the only difference being the epoch setting, which is adjusted to 100 to ensure a fair comparison with RSP. We evaluate the model in the same setting, which is detailed in the Supplementary Material.

\textbf{Datasets.} MillionAID is a large-scale dataset in the remote sensing (RS) domain, comprising 1,000,848 non-overlapping scenes. Notably, MillionAID is RGB-based, making it more compatible with existing deep models developed in the natural image domain. The MillionAID dataset comprises 51 categories organized in a hierarchical tree structure, with 51 leaves distributed across 28 parent nodes at the second level. 
The images vary in size from 110 $\times$ 110 to 31,672 $\times$ 31,672. 
We utilize the same dataset split as RSP, which randomly chooses 1,000 images in each category to form the validation set of 51,000 images, with the remaining 949,848 images used for training.

\textbf{Results.} Results for MillionAID aerial scene recognition is summarized in Table~\ref{tab:millionaid}. We mainly compare LeMeViT-Small with the representative Swin Transformer and the SOTA ViTAE. Compared to Swin-Tiny, LeMeViT-Small achieves 1.25 $\times$ faster inference speed, and its accuracy trained for 100 epochs surpasses Swin-Tiny trained for 300 epochs by 0.41\%. In comparison to ViTAE, our model achieves nearly a 1.96$\times$ speedup in inference, a 1.47 $\times$ speedup in training, and even a slight improvement in accuracy by 0.03\%. Other LeMeViT variants also show competitive efficiency and performance.

\subsection{Remote Sensing Downstream Tasks} \label{downstream}
To validate the transferability of LeMeViT to dense prediction tasks, we apply the pre-trained models from Sec.~\ref{millionaid} to various remote sensing downstream tasks, including object detection, semantic segmentation, and change detection. Following RSP, we fine-tune the models using appropriate methods and settings for each task. Specific training recipes are provided in Supplementary Material.

\textbf{Object Detection.} 
Aerial object detection involves detecting oriented bounding boxes (OBB) instead of the typical horizontal bounding boxes (HBB) used in conventional natural image tasks. We conduct aerial object detection experiments using the DOTA dataset~\cite{Xia_2018_dota}, which is the most famous large-scale dataset for OBB detection. It contains 2,806 images with sizes ranging from 800 $\times$ 800 to 4,000 $\times$ 4,000, encompassing 188,282 instances across 15 categories. We use Oriented-RCNN \cite{xie_oriented_2021} as the OBB detection head. 
We train models on the merged training set and validation set of DOTA datasets and evaluate them on the testing set, which is only accessible on the evaluation server. We use the mean average precision (mAP) of all categories as the metric.

\textbf{Semantic Segmentation.} Aerial semantic segmentation refers to pixel-level classification of the aerial scene. We use the ISPRS Potsdam\footnote{\url{https://www.isprs.org/education/benchmarks/UrbanSemLab/2d-sem-label-potsdam.aspx}} dataset as the benchmark for this task. 
This dataset has 38 images, each with an average size of 6,000 $\times$ 6,000 pixels. These images are cropped into 512 $\times$ 512 patches with a stride of 384, and they contain six categories: impervious surface, building, low vegetation, tree, car, and clutter. The dataset is divided into training and testing sets, with 24 and 14 images respectively. We use UperNet~\cite{xiao_unified_2018} as the segmentation framework. 
The overall accuracy (OA) and mean F1 score (mF1) are used for evaluation.

\begin{table*}[t]
\centering
\begin{minipage}[t]{.8\textwidth}
\centering
\renewcommand{\arraystretch}{1}
\setlength{\tabcolsep}{6pt}
\begin{tabular}{l!{\vrule width1.2pt}cc!{\vrule width1.2pt}ccc!{\vrule width1.2pt}cc}
   \bottomrule[1.5pt]
\multirow{2}{*}{Backbone} & \multicolumn{2}{c!{\vrule width1.2pt}}{Object Detection}                 & \multicolumn{3}{c!{\vrule width1.2pt}}{Semantic Segmentation}                             & \multicolumn{2}{c}{Change Detection}                 \\ \cline{2-8} 
                          & \multicolumn{1}{c|}{mAP $\uparrow$}            & MACs $\downarrow$           & OA $\uparrow$            & \multicolumn{1}{c|}{mF1 $\uparrow$}            & MACs $\downarrow$           & \multicolumn{1}{c|}{mF1 $\uparrow$}            & MACs $\downarrow$          \\ \Xhline{1pt}
Swin-tiny                 & \multicolumn{1}{c|}{76.50}          & 215.68          & 90.78          & \multicolumn{1}{c|}{90.03}          & 234.79          & \multicolumn{1}{c|}{95.21}          & 15.63          \\
ViTAE-v2-small            & \multicolumn{1}{c|}{77.72}          & 234.82          & 91.21          & \multicolumn{1}{c|}{90.64}          & 238.28          & \multicolumn{1}{c|}{96.81}          & 15.94          \\ \hline
LeMeViT-tiny             & \multicolumn{1}{c|}{76.63}          & \textbf{154.12} & 91.03          & \multicolumn{1}{c|}{90.55}          & \textbf{217.88} & \multicolumn{1}{c|}{95.56}          & \textbf{5.75} \\
LeMeViT-small             & \multicolumn{1}{c|}{77.58}          & 193.91 & 91.23          & \multicolumn{1}{c|}{90.62}          & 228.16 & \multicolumn{1}{c|}{96.64}          & 10.71 \\
LeMeViT-base              & \multicolumn{1}{c|}{\textbf{78.00}} & 335.53          & \textbf{91.35} & \multicolumn{1}{c|}{\textbf{90.85}} & 263.75          & \multicolumn{1}{c|}{\textbf{97.32}} & 28.47          \\ \toprule[1.5pt]
\end{tabular}
   \caption{ Comparison on three downstream tasks. More details are provided in Supplementary Material.
   \label{tab:downstream}}
\end{minipage}
\end{table*}

\begin{figure*}[t]
{
\centering
\includegraphics[width=0.86\textwidth]{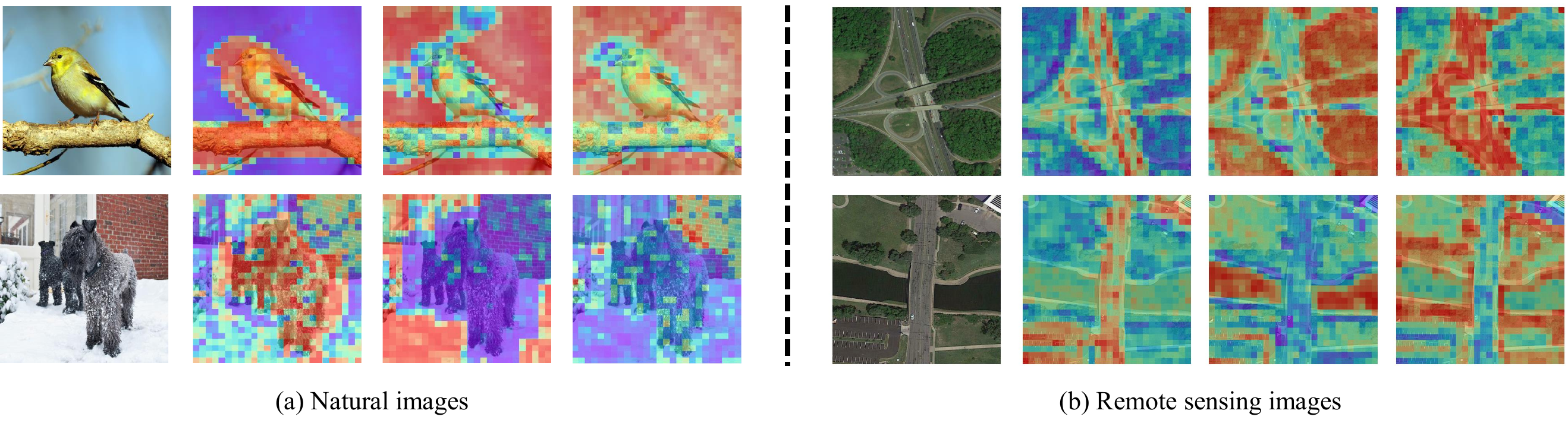}
\caption{Visualization of the attention maps between three meta tokens in the last layer and image tokens. \textbf{(a)} illustrates the attention maps on natural images, while \textbf{(b)} illustrates attention maps on remote sensing images.
\label{fig:visualization}
}}
\end{figure*}

\textbf{Change Detection.} Aerial change detection involves binary classification to pixel-wise label the dissimilarities between two images captured in the same scene at different timestamps. We use the pre-processed CDD dataset~\cite{lebedev_change_2018} to evaluate models on this task. The image pairs are cropped into a sequence of 256 $\times$ 256 patches, and the sizes of the training, validation, and testing sets are 10,000/3,000/3,000, respectively. We adopt the BIT~\cite{chen_remote_2022} framework for change detection. 
We report the mean F1 score (mF1) on the testing set.

\textbf{Results.} The results of LeMeViT-Small and other models are presented in Table \ref{tab:downstream}. Since the inference speed depends on many factors, such as the number of instances in an image for object detection, we mainly report MACs as the efficiency metric. Further details and results of other models can be found in the Supplementary Material. In the object detection task, LeMeViT-Small exhibits only a 0.14\% mAP loss but has 20\% less computations compared to ViTAEv2-Small, and outperforms Swin-Tiny in both detection accuracy and computational efficiency. For semantic segmentation, there is almost no difference in OA and mF1 between LeMeViT and ViTAE models. In change detection, the mF1 score of our model is inferior to ViTAEv2-Small but comparable with Swin-Tiny, but they require almost 50\% more computations.

\subsection{Ablation Study}  \label{ablation}
In the ablation studies, we primarily investigate the impact of the length of meta tokens. The results are shown in the Table~\ref{tab:ablation}. An interesting conclusion can be drawn that the length of meta tokens has a marginal impact on the performance. With lengths of 64, 32, 16, and 8, the accuracy is almost the same. This further confirms the redundancy in images and the vanilla attention calculation, suggesting the motivation of using a smaller number of meta tokens to represent the dense image tokens. Finally, considering both efficiency and accuracy, we choose 16 as the default setting of meta token length.

More ablation studies are conducted to validate the effectiveness of components, \textit{i.e.}, cross attention block, meta token stem and token fusion method in the final layer. Results provided in Supplementary Material demonstrate these designs increase the accuracy.

\begin{table}[h]
\centering
\renewcommand{\arraystretch}{1.1}
\setlength{\tabcolsep}{4.2pt}
\begin{tabular}{c|cc|c}
   \bottomrule[1.5pt]
Token Length                     & \textbf{\begin{tabular}[c]{@{}c@{}} Inference \\ Throughputs $\uparrow$ \end{tabular}} 

& \textbf{MACs} $\downarrow$ & \textbf{\begin{tabular}[c]{@{}c@{}}MillionAID \\ Acc@1 $\uparrow$ \end{tabular}} \\ \hline
64          & 3268.48                                                               & 4.39          & 98.96                                                                \\
32          & 3481.31                                                               & 3.95          & 98.97                                                                \\
\textbf{16} & 3609.64                                                               & 3.74          & \textbf{99.00 }                                                         \\
8           & 3639.84                                                               & 3.63          & 98.96                                                                \\ \toprule[1.5pt]
\end{tabular}
   \caption{ Results of different settings of meta token length in LeMeViT-small. 
   \label{tab:ablation}
   }
\end{table}

\subsection{Visualization} \label{visualization}
To gain a deeper understanding of how meta tokens work, we visualize the cross-attention maps between meta tokens and image tokens in the last block of the DCA, as illustrated in Fig.~\ref{fig:visualization}. We visualize both natural images and remote sensing images. The cross-attention in natural images (Fig.~\ref{fig:visualization}(a)) reveals that the learned meta tokens can well attend to semantic parts of images, \textit{i.e.}, foreground objects, leading to a better object representation and effective information exchange with image tokens, which contribute to the improved classification accuracy. 
Fig.~\ref{fig:visualization}(b) shows the cross-attention for different meta tokens, providing a clear indication that different meta tokens are responsible for different semantic parts of images, \textit{e.g.}, roads, grasslands, and forests. The results imply that the meta tokens can learn effective representations by aggregating important semantic regions in images. The visualization offers a clear way to explain how meta tokens function, enhancing the interpretability of LeMeViT.

\section{Conclusion}
This paper introduces LeMeViT, a novel Vision Transformer architecture designed to efficiently address computational bottlenecks in traditional attention layers. Inspired by the spatial redundancy in images, particularly in remote sensing images, we suggest learning meta tokens to represent dense image tokens. To enhance computational efficiency, we replace the original self-attention mechanism with a Dual Cross-Attention, promoting information exchange between meta tokens and image tokens. LeMeViT is versatile, supporting image classification, scene recognition and diverse dense prediction tasks. Experimental results on different public benchmarks show LeMeViT's superior balance between efficiency and performance.

\newpage
\section*{Acknowledgments}
This work was supported in part by the National Natural
Science Foundation of China under Grants 62225113, 62271357, the National Key Research and Development Program of China 2023YFC2705700, the Natural Science
Foundation of Hubei Province under Grants 2023BAB072,
and the Fundamental Research Funds for the Central Universities
under Grants 2042023kf0134.

\bibliographystyle{named}
\bibliography{refs/ms}

\end{document}


\newpage

\appendix
\section{Implementation Details}

\subsection{Image Classification on ImageNet-1K} \label{supp:classification}
For image classification on ImageNet-1K~\cite{deng2009imagenet}, all models are trained for 300 epochs on training set from scratch and tested on validation set. For all training settings, the input image resolution is set to $224 \times 224$. The training settings most follow DeiT~\cite{touvron2022deit}, which has been widely used. We train our models on four Nvidia RTX 4090 GPUs and test on one with automatic mixed precision (AMP). We use the latest version of TIMM~\cite{rw2019timm} for testing and benchmarking.

\textbf{Training.} 
AdamW~\cite{loshchilov2017decoupled} is employed as the optimizer, and cosine decay learning rate scheduler is applied to adjust learning rate. 
A momentum of 0.9, a betas of (0.9,0.999), a weight decay of 0.05 are used.
The initial learning rate is set to 5.0 $e{-4}$, and warmup\_lr, min\_lr is set to 5.0 $e{-6}$. All the learning rates are linearly scaled by global batch size, where they are multiplied by the batch size and the number of devices, and then divided by a scale constant of 512. The learning rate is first warmed up linearly for 5 epochs from the warmup\_lr to the initial learning rate, and then dropped to min\_lr using the cosine scheduler. Label smoothing cross-entropy~\cite{szegedy2016rethinking} with smoothing of 0.1 is employed as the loss function. We use most data augment and regularization techniques, \textit{i.e.}, "rand-m9-mstd0.5-inc1" RandAugment~\cite{cubuk2020randaugment} , Mixup~\cite{zhang2017mixup} of 0.8 with prob of 1.0, Cutmix~\cite{yun2019cutmix} of 1.0, ColorJitter~\cite{krizhevsky2012imagenet} of 0.4, Horizontal Flip of 0.5, stochastic depth~\cite{huang2016deep} of 0.1 and repeated augmentation in model training. We set different batch sizes, \textit{i.e.}, 400, 256, and 64, for our Tiny, Small, and Base model variants, respectively.  During testing, we only utilize resizing to transform the image resolution to 224 $\times$ 224, without employing any additional data augmentation.

\textbf{Benchmarking.} 
We use calflops~\cite{calflops} to calculate the parameter count and MACs for all the involved models. To test throughput, experiments are conducted on an RTX 4090 GPU. During testing, the maximum batch size is set to 256, gradually halving if there was insufficient memory. The models are run with Automatic Mixed Precision (AMP) for 100 iterations, and the final average throughputs are reported as the result.

\subsection{Remote Sensing Scene Recognition} \label{supp:recognition}
In the experiments of remote sensing scene recognition, the majority of settings follow the aforementioned image classification in Sec.\ref{supp:classification}. The only difference is that the training epochs are set to 100, as the MillionAID~\cite{long_creating_2021} dataset converges more quickly and for fair comparison. Additionally, we monitor GPU usage to assess memory consumption. Notably,  MillionAID dataset has only 51 categories, much fewer than ImageNet-1K's 1000 categories, leading to slight differences in models' parameter counts and throughputs.

\subsection{Aerial Object Detection}
In aerial object detection, we conduct a series of experiments using the Oriented-RCNN~\cite{xie_oriented_2021} as the detection head on the DOTA dataset~\cite{Xia_2018_dota}. We train models on the merged training and validation sets and evaluate them on the testing set, which is only accessible on the evaluation server. The models are trained for 12 epochs on four RTX 4090 GPUs, with a batch size of 8 per GPU. AdamW with a betas of (0.9,0.999) and a weight decay of 0.05 is used as the optimizer, and the initial learning rate is set to 0.001. StepLR is employed for learning rate adjustment, dividing the learning rate by 10 at the 8th and 11th epochs. A linear warm-up of the learning rate is performed for the first 500 iterations. Following Oriented-RCNN, the raw images are sampled and cropped into size of 1024 $\times$ 1024, with a stride of 824. Data augmentation strategies include Random Rotation, Horizontal Flip, and Vertical Flip, all with a probability set to 0.5.

\subsection{Aerial Semantic Segmentation}

In the aerial semantic segmentation task, we use UperNet~\cite{xiao_unified_2018} as our segmentation head, which is implemented by mmsegmentation~\cite{mmseg2020}. We conduct experiments on the ISPRS Potsdam\footnote{\url{https://www.isprs.org/education/benchmarks/UrbanSemLab/2d-sem-label-potsdam.aspx}}, which contains six classes, including impervious surface, building, low vegetation, tree, car, and clutter, where the clutter category is ignored during both training and evaluation.
The images are cropped into 512 $\times$ 512 patches with a stride of 384, and only augmented by Random Flip with probability of 0.5. We use AdamW with a betas of (0.9,0.999) and a weight decay of 0.01 as the optimizer, and 0.0002 as the initial learning rate. The learning rate is adjusted by polynomial learning rate
policy with warm restart~\cite{mishra2019polynomial} (PolyLR). The power of PolyLR is set to 1.0, and warm-up iterations is set to 1500. The models are trained for 8000 iterations using four RTX 4090 GPUs, employing a batch size of 10 per GPU. 

\subsection{Aerial Change Detection}
We employ the BIT~\cite{chen_remote_2022} framework for the change detection task. We conduct experiments on the pre-processed CDD dataset~\cite{lebedev_change_2018}. For this task, we utilize a single RTX 4090 for both training and testing, without using Automatic Mixed Precision (AMP). During training, the batch size is set to 40, and the number of epochs is set to 200. AdamW is utilized as the optimizer with a betas of (0.9,0.999) and a weight decay of 0.01. The learning rate is adjusted by the linear learning rate policy (LinearLR). We use binary cross-entropy as the loss function.

\section{Experimental Results Details}

In Sec.4.3 of the main paper, we present a summary of the experimental results for various downstream tasks. More detailed experimental results are provided in this section, encompassing all versions of our model, along with results from other models. The results are further detailed by category.

\begin{table*}[h]
\renewcommand{\arraystretch}{1.2}
\setlength{\tabcolsep}{3pt}
\setlength{\belowcaptionskip}{7pt}
\centering
\footnotesize
\begin{threeparttable}
\begin{tabular}{l|c|ccccccccccccccc}
\bottomrule
\multirow{2}{*}{Backbone} & \multirow{2}{*}{mAP} & \multicolumn{15}{c}{AP of each category}                                                                                                                                                                                                                     \\ \cline{3-17} 
                          &                      & Plane          & BD             & Bridge         & GTF            & SV             & LV             & Ship           & TC             & BC             & ST             & SBF            & RA             & Harbor         & SP             & HC             \\ \hline
ResNet-50                 & 76.50                & \textbf{89.78} & 81.88          & 54.39          & 70.91          & 78.67          & 83.01          & \textbf{88.17} & 90.84          & 86.17          & \textbf{85.72} & 62.22          & 67.45          & 73.99          & 72.21          & 62.22          \\
Swin-tiny                 & 76.12                & 89.54          & 79.74          & 52.91          & 74.50          & 78.96          & 84.02          & 87.83          & 90.86          & 85.90          & 84.84          & 62.90          & 67.33          & 74.45          & 70.61          & 57.36          \\
ViTAE-small               & 77.72                & 89.66          & 83.04          & 55.85          & \textbf{75.16} & \textbf{79.95} & 84.34          & 88.04          & \textbf{90.90}          & \textbf{88.17} & 85.58          & 62.64          & 70.60 & 76.77          & 67.15          & 67.89          \\ \hline
LeMeViT-tiny              & 76.63                & 89.34 & 80.80 & 52.96 & 72.64 & 78.99 & 83.75 & 87.93 & 90.87 & 87.01 & 85.08 & \textbf{63.61} & \textbf{71.15} & 74.74 & 69.65 & 61.00          \\
LeMeViT-small             & 77.58                & 89.42 & 83.00 & 55.51 & 72.38 & 78.82 & \textbf{84.60} & 87.75 & 90.89 & 87.13 & 85.37 & 58.84 & 70.60 & 75.98 & \textbf{78.57} & 64.88          \\
LeMeViT-base              & \textbf{78.00}       & 89.49          & \textbf{83.56} & \textbf{56.10} & 74.02          & 79.67          & 84.50 & 87.98          & 90.89          & 87.78          & 84.99          & 62.92          & 70.19          & \textbf{76.86} & 71.33          & \textbf{69.78} \\ \toprule
\end{tabular}
\begin{tablenotes}
        \footnotesize
        \item[*] BD: baseball diamond; GTF: ground track field; SV: small vehicle; LV: large vehicle; TC: tennis court; BC: basketball court, ST: storage tank; SBF: soccer ball field; RA: roundabout; SP: swimming pool; HC: helicopter.
      \end{tablenotes}
    \end{threeparttable}
   \caption{Results of Object Detection on DOTA. 
   \label{supp:object detection}
   }
\vspace{-10pt}
\end{table*}

\begin{table*}[]
\footnotesize
\renewcommand{\arraystretch}{1.2}
\setlength{\tabcolsep}{5pt}
\begin{minipage}[t]{0.7\textwidth}
\centering
\begin{tabular}{l|c|c|ccccc}
\hline
\multirow{2}{*}{Backbone} & \multirow{2}{*}{OA} & \multirow{2}{*}{mF1} & \multicolumn{5}{c}{F1 score of each category}                                      \\ \cline{4-8} 
                          &                     &                      & Imper.Surf             & Building             & Low.Veg             & Tree           & Car            \\ \hline
ResNet-50                 & 90.61               & 89.94                & 92.42          & 96.15          & 85.75          & 85.49          & 89.87          \\
Swin-tiny                 & 90.78               & 90.03                & 92.65          & 96.35          & 86.02          & 85.39          & 89.75          \\
ViTAE-small               & 91.21               & 90.64                & 93.05          & 96.62          & 86.62          & \textbf{85.89} & 91.01          \\ \hline
LeMeViT-tiny              & 91.03               & 90.55                & 92.87          & 96.41          & 86.49          & 85.65          & 91.32          \\
LeMeViT-small             & 91.23               & 90.62                & 92.96          & 96.58          & 86.55          & 85.65          & 91.36          \\
LeMeViT-base              & \textbf{91.35}      & \textbf{90.85}       & \textbf{93.15} & \textbf{96.78} & \textbf{86.93} & 85.80          & \textbf{94.22} \\ \hline
\end{tabular}
   \caption{Results of Semantic Segmentation on ISPRS Potsdam.
   \label{supp:semantic segmentation}}
\end{minipage}  \begin{minipage}[t]{0.3\textwidth}
\centering
\setlength{\belowcaptionskip}{5pt}
\begin{tabular}{l|c}
\hline
Backbone      & F1 score       \\ \hline
ResNet-50     & 96.00          \\
Swin-tiny     & 95.21          \\
ViTAE-small   & 96.81          \\ \hline
LeMeViT-tiny  & 95.56          \\
LeMeViT-small & 96.64          \\
LeMeViT-base  & \textbf{97.32} \\ \hline
\end{tabular}
   \caption{Results of Change Detection on CDD.
   \label{supp:change detection}}
\end{minipage}
\vspace{-14pt}
\end{table*}

Specifically, in Table~\ref{supp:object detection}, Table~\ref{supp:semantic segmentation}, and Table~\ref{supp:change detection}, we present detailed results for object detection, semantic segmentation, and change detection, respectively. Table~\ref{supp:params and macs} presents the models' parameter counts and MACs for different downstream tasks, which is tested using calflops. The showcased models include ResNet-50~\cite{he2016deep}, Swin Transformer~\cite{liu_swin_2021}, ViTAE~\cite{xu_vitae_2021}, and three versions of LeMeViT (Tiny, Small, and Base). Some data is sourced from RSP~\cite{wang_empirical_2023}, and for a fair comparison, we leverage results of models using remote sensing pre-training on MillionAID dataset. 

In the table, we use abbreviations to represent these classes. We report results of mean Average Precision (mAP) and Average Precision (AP) for each class.
The Potsdam dataset comprises 6 classes: impervious surface, building, low vegetation, tree, car, and clutter (with clutter being disregarded in our evaluation). Our reported metrics include Overall Accuracy (OA), mean F1 score (mF1), and F1 score for each class.
For change detection, which is a binary classification task, we present F1 scores as the metrics.

\begin{table}[h]
\vspace{-4pt}
\renewcommand{\arraystretch}{1.1}
\setlength{\tabcolsep}{3pt}
\centering
\begin{tabular}{l|cc}
\hline
                       & with DCA (img/sec) & with SA (img/sec) \\ \hline
\textbf{LeMeViT-tiny}  & 5460.58            & 4048.07           \\
\textbf{LeMeViT-small} & 3608.12            & 2625.74           \\
\textbf{LeMeViT-base}  & 1482.70             & 1061.57           \\ \hline
\end{tabular}%

   \caption{\fontsize{3bp}{2bp} Comparison of DCA and SA.
   \label{tab:dca}
}
\vspace{13pt}

\begin{tabular}{l|c}
\hline
Model Design              & ImageNet-1K Acc@1 \\ \hline
LeMeViT-small                    & \textbf{78.37}    \\
w/o cross attention block & 78.10             \\
w/o meta token stem       & 78.19             \\
w/o meta token pooling    & 78.29             \\ \hline
\end{tabular}

   \caption{Ablation studies with different model designs.
   \label{supp:ablation}}
\end{table}

\section{More Ablation Studies}
To explore the roles of different modules in the architecture, we conduct ablation studies on various components, focusing on the Cross Attention block, meta token stem, and token fusion for the classification task, respectively. We conduct experiments with classification tasks on ImageNet-1K dataset, using LeMeViT-Small as the baseline model. The experimental settings are similar to those described in Sec.~\ref{supp:classification}, with the only difference being the epochs set to 100 to expedite the experiments for preliminary performance evaluation. 

For the Cross Attention block, we remove it from LeMeViT-Small and test its accuracy. For the meta token's Stem layer, we similarly exclude it and adjust the initial dimension of meta tokens to match the output dimension of the Stem for experimentation. In the original LeMeViT, the output image tokens and meta tokens are fused for classification, by separately applying global pooling and then adding them together. In our ablation experiments, we utilize only the image tokens for classification, without meta token pooling.

The experimental results are presented in the Table~\ref{supp:ablation}, demonstrating that these designs do improve the model's performance. When these designs are removed, the model exhibits varying degrees of accuracy decrease on the ImageNet-1K dataset.

\begin{table*}[t]
\centering
\begin{tabular}{l|cc|cc|cc}
\hline
\multirow{2}{*}{Backbone} & \multicolumn{2}{c|}{Object Detection} & \multicolumn{2}{c|}{Semantic Segmentation} & \multicolumn{2}{c}{Change Detection} \\ \cline{2-7} 
                          & Params            & MACs              & Params               & MACs                & Params            & MACs             \\ \hline
ResNet-50                 & 41.14             & 211.43            & 66.39                & 235.49              & 24.44             & 12.53            \\
Swin-tiny                 & 44.46             & 215.68            & 59.83                & 234.87              & 28.27             & 15.63            \\
ViTAE-small               & 35.02             & 234.82            & 49.05                & 238.58              & 19.56             & 15.94            \\ \hline
LeMeViT-tiny              & 25.37             & 154.12            & 37.05                & 217.88              & 8.56              & 5.75             \\
LeMeViT-small             & 33.15             & 193.91            & 45.59                & 228.16              & 16.75             & 10.71            \\
LeMeViT-base              & 69.76             & 335.53            & 83.19                & 263.75              & 53.34             & 28.47            \\ \hline
\end{tabular}
   \caption{Model parameters and MACs across various downstream tasks.
   \label{supp:params and macs}}
\end{table*}

\begin{table*}[h]
\centering
\renewcommand{\arraystretch}{1.1}
\setlength{\tabcolsep}{1.8pt}
\begin{tabular}{l|ccccccccc}
\hline
                       & \multicolumn{9}{c}{RetinaNet $1\times$ schedule}                                              \\ \cline{2-10} 
              & $mAP$  & $AP_{50}$ & $AP_{75}$ & $AP_S$  & $AP_M$  & \multicolumn{1}{c|}{$AP_L$}  & Params & MACs   & Infer  \\ \hline
Swin-tiny     & 41.5 & 62.1 & 44.2 & 25.1 & 44.9 & \multicolumn{1}{c|}{55.5} &      38.16  &   232.11     &    35.1  \\
\textbf{LeMeViT-small} & \textbf{43.1} & \textbf{64.3} & \textbf{46.2} & \textbf{26.3} & \textbf{46.5} & \multicolumn{1}{c|}{\textbf{56.7}} & \textbf{25.96}  & \textbf{212.34} & \textbf{44.8} \\ \hline
\end{tabular}
\vspace{7pt}

\renewcommand{\arraystretch}{1.1}
\setlength{\tabcolsep}{1.2pt}
\begin{tabular}{l|ccccccccc}
\hline
                       & \multicolumn{9}{c}{Mask-RCNN $1\times$ schedule}                                                      \\ \cline{2-10} 
              & $mAP^b$  & $AP^b_{50}$ & $AP^b_{75}$ & $mAP^m$  & $AP^m_{50}$ & \multicolumn{1}{c|}{$AP^m_{75}$} & Params & MACs   & Infer \\ \hline
Swin-tiny     & 42.2 & 64.6 & 46.2 & 39.1 & 61.6 & \multicolumn{1}{c|}{42.0}   & 47.49  &     253.43   &     23.7     \\
\textbf{LeMeViT-small} & \textbf{44.4} & \textbf{66.9} & \textbf{48.1} & \textbf{39.9} & \textbf{63.7} & \multicolumn{1}{c|}{\textbf{43.5}} & \textbf{36.18}  & \textbf{228.05} & \textbf{28.5}         \\ \hline
\end{tabular}
   \caption{Results of semantic segmentation on COCO.
   \label{supp:coco}}
\end{table*}

\section{Results on COCO datasets}

We perform object detection and instance segmentation experiments on COCO dataset by employing RetinaNet and Mask-RCNN as detection and segmentation heads. We only compare LeMeViT and Swin Transformer. The Table \ref{supp:coco} below shows that LeMeViT outperforms Swin Transformer in precision, inference speed, and parameter count.

\bibliographystyle{named}
\bibliography{refs/supplement}